\documentclass[letterpaper, 10 pt, conference]{ieeeconf}  
\IEEEoverridecommandlockouts                              
\usepackage[pdftex]{graphicx}
\usepackage{url}
\usepackage{here}
\usepackage{bm}
\usepackage[subrefformat=parens]{subcaption}
\usepackage{threeparttable}
\title{\LARGE \bf
Generation Drawing/Grinding Trajectoy Based on Hierarchical CVAE
}

\author{Masahiro Aita, Keito Sugawara, Sho Sakaino, \textit{Senior Member, IEEE} and Toshiaki Tsuji, \textit{Senior Member, IEEE}
\thanks{Masahiro Aita, Keito Sugawara and Toshiaki Tsuji ith the Department of Electrical, and
Electronic Systems, Saitama University, Saitama 338-8570, Japan 
(e-mail: m.aita.378@ms.saitama-u.ac.jp, tsuji@ees.saitama-u.ac.jp)}%
\thanks{Sho Sakaino is with the Graduate School of Systems and Information Engineering, University of Tsukuba, 1-1-1 Tennodai, Tsukuba, Ibaraki 305-8577, Japan (e-mail: sakaino@iit.tsukuba.ac.jp).
}
}
\begin{document}

\maketitle
\thispagestyle{empty}
\pagestyle{empty}

\begin{abstract}

        In this study, we propose a method to model the local and global features of the drawing/grinding trajectory with hierarchical Variational Autoencoders (VAEs). By combining two separately trained VAE models in a hierarchical structure, it is possible to generate trajectories with high reproducibility for both local and global features. The hierarchical generation network enables the generation of higher-order trajectories with a relatively small amount of training data. The simulation and experimental results demonstrate the generalization performance of the proposed method. In addition, we confirmed that it is possible to generate new trajectories, which have never been learned in the past, by changing the combination of the learned models. 

\end{abstract}

\begin{keywords}
        Learning from demonstration, imitation learning, variational autoencoder, hierarchical
\end{keywords}
\section{INTRODUCTION}
\PARstart{I}{n} recent years, generative networks technology has been advancing rapidly. In particular, the development of Generative Adversarial Network (GAN)~\cite{goodfellow} has enabled the generation of images and sounds too realistic for humans to distinguish. This deep fake technology generates new data that does not exist, which has resulted in a significant social impact and various problems owing to its performance.  

In the field of robot control, attempts to apply machine learning to the generation of motions are advancing. Learning from demonstration (LfD) is widely conducted as a methodology to imitate human motions and generate new ones~\cite{Schaal1997}\cite{Argall}\cite{lfd}. 
The Gaussian Mixture Model (GMM)~\cite{gmm} , Hidden Markov Model, (HMM)~\cite{hmm}, and Neural Network (NN)~\cite{rnn} have been often employed as learning models. In addition, there have been several applications of autoencoders, a type of generative network~\cite{PMLR}. One of the advantages of using generative networks is that they can generate new behaviors, different from those learned in the past. For example, Kutsuzawa \textit{et al}. showed that the Sequence-to-Sequence (Seq2Seq) model, which is a type of autoencoder, could generate new actions, such as not dropping the pancake by learning the new action of flipping the pancake~\cite{kutsuzawa}.

There are many studies on Variational Autoencoders (VAEs)~\cite{vae}, extending the scope to the generation of time-series positional information without limiting it to robots. VAEs consist of two models: an encoder, which transforms the input into low-dimensional latent variables, and a decoder, which generates the output from the latent variables with the decoder functioning as a generative network. Its extension, conditional VAE (CVAE)~\cite{cvae}, generates data associated with labels by assigning labels to the decoder along with the latent variables, and it is used for generating images~\cite{face} and music~\cite{music} in addition to anomaly detection~\cite{AD}. Ha \textit{et al}. used a Sketch-RNN composed of a VAE to generate images that preserve the writing order by capturing sketch images as images composed of time series vectors~\cite{sketch}. Zhang \textit{et al}. attempted to generate vectorized Chinese characters by modeling the pen movements of writing the Chinese characters using an RNN~\cite{kanji}. In all these studies, multiple actions were generated from a single model; however, the actions had to be prepared as training data in advance. To generate a new action, it is necessary to add training data corresponding to the action and relearn it.

When the task becomes a higher-order task, it becomes difficult to prepare all the characteristic behaviors in the training data in advance. In the field of reinforcement learning, it is known that hierarchical models~\cite{hierarchical1}\cite{hierarchical2} are effective in dealing with this limitation. The literature~\cite{Hierarchical} shows that the amount of training data required can be reduced through hierarchical models. Factory automation has been developed by reproducing human instruction data with high reproducibility and establishing mass production technology; however, for expansion to high-mix low-volume production, technology to generate a large number of actions from a limited number of instruction patterns is necessary. In this study, we propose a multistage CVAE that applies the concept of hierarchical learning to robot motion generation to meet this need.

\begin{figure*}[t]
    \begin{minipage}[b]{0.5\linewidth}
        \centering
            \includegraphics[width=80mm]{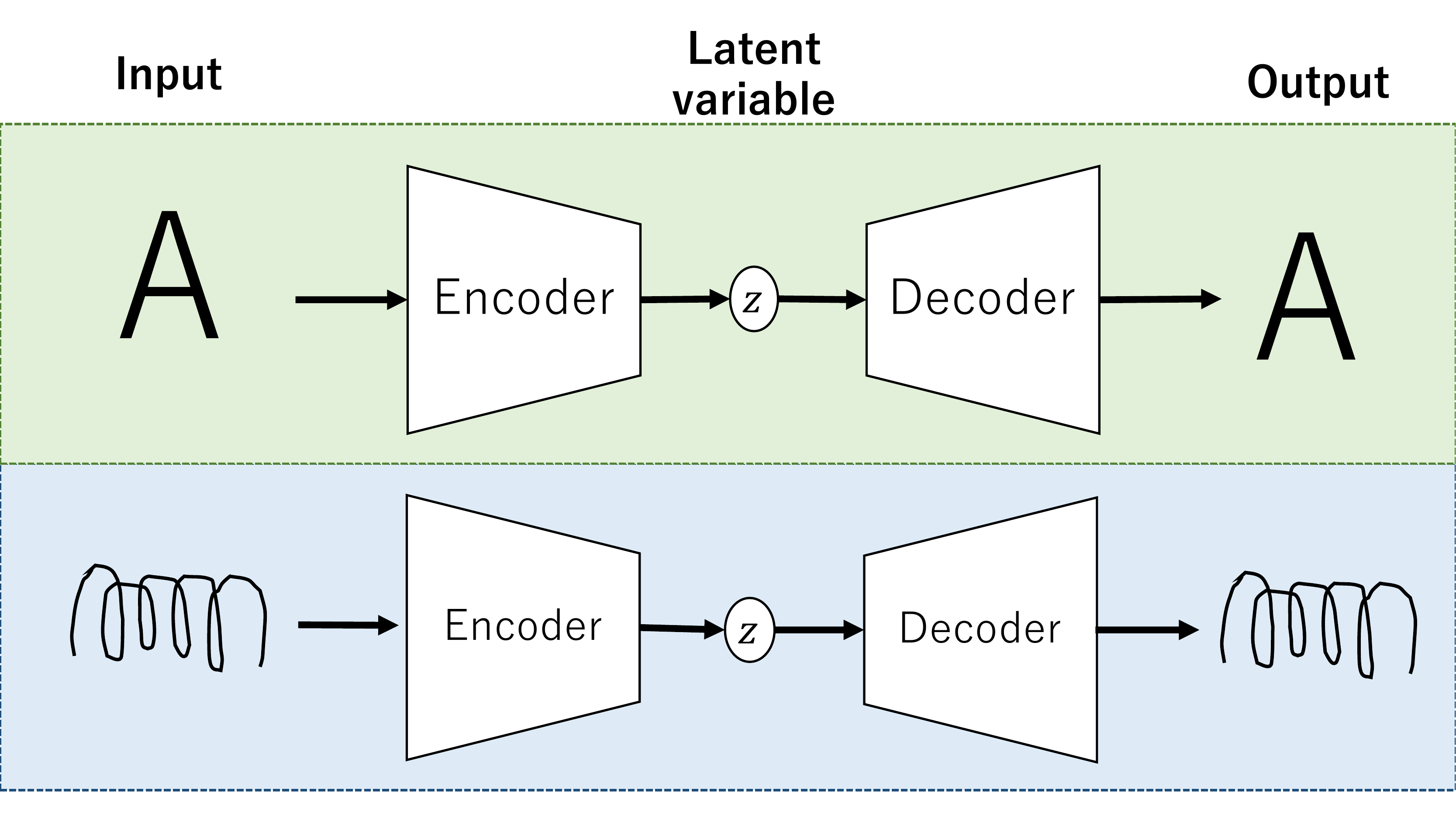}
            \subcaption{Original VAE}%
            \label{fig:vae}
    \end{minipage}
    \begin{minipage}[b]{0.5\linewidth}
        \centering
            \includegraphics[width=80mm]{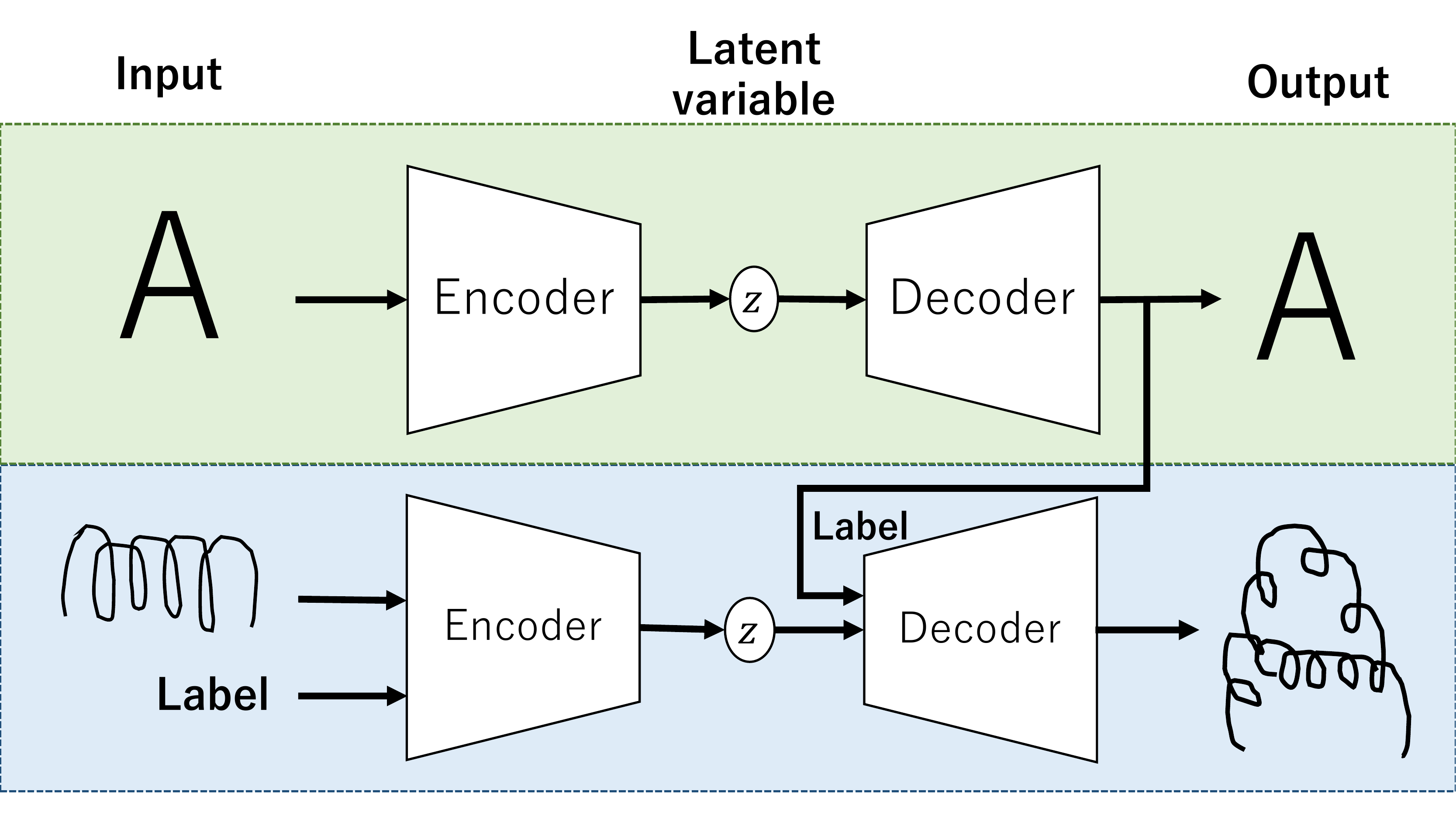}
            \subcaption{Proposed method}
            \label{fig:proposed}
    \end{minipage}
    \caption{\leftline{ Comparison of VAE and proposed method.}}
    \label{fig:proposedmethod}
\end{figure*}

The proposed concept is illustrated in Fig.~\ref{fig:proposedmethod}. Conventional techniques learn the trajectories from position and force profiles. The Seq2Seq model, which constitutes an autoencoder in LSTM, can deal with both short-term and long-term trajectories; however, it can only learn the motion contained in the training data. In the proposed method, the CVAE is multistaged, and the output of the upper decoder (the position and force of the endpoint of a single stroke trajectory) is the input as the label of the CVAE of the lower decoder. Therefore, the proposed method can represent higher-order trajectories and generate more trajectory patterns by changing the combination of decoders.

The primary contribution of this study is to show that the generation of higher-order trajectories can be achieved through multistage generation networks with the same structure. The generated network of an image has a complex structure and is multileveled to facilitate higher-order expressiveness. On the contrary, the trajectory of a robot is represented by a 12-dimensional time series, including position and force, and does not require a model as complex as that of an image. MTRNN~\cite{tani} is known as a model for learning time series information with hierarchical RNNs. In addition, Demiris \textit{et al}. have shown that the more layers of Hierarchical behavioral repository, a kind of autoencoder, are used, the higher-order trajectories can be represented~\cite{demiris}. In this study, CVAE is multileveled and both endpoints of a stroke are passed from the upper layer to the lower layer. The hierarchical structure is connected by explicit variables, which is advantageous because the hierarchical structure can be freely rearranged.

\section{METHODS}
In this section, we describe the structure of the proposed hierarchical CVAE method and the training method of the model. In this study, we used the trajectories of human writing as the learning data. When drawing a thick line during grinding or sketching, the robot should move back and forth in a fine reciprocal motion of line width, and the repetition of this motion makes one long line. This localized reciprocating motion, which changes the textures of the line is called touch, and the global motion drawn by the repetition of touch is called the stroke order. 

\subsection{Structure and Learning of Hierarchical CVAE}
\begin{figure}[t]
    \centering
        \includegraphics[width=50mm]{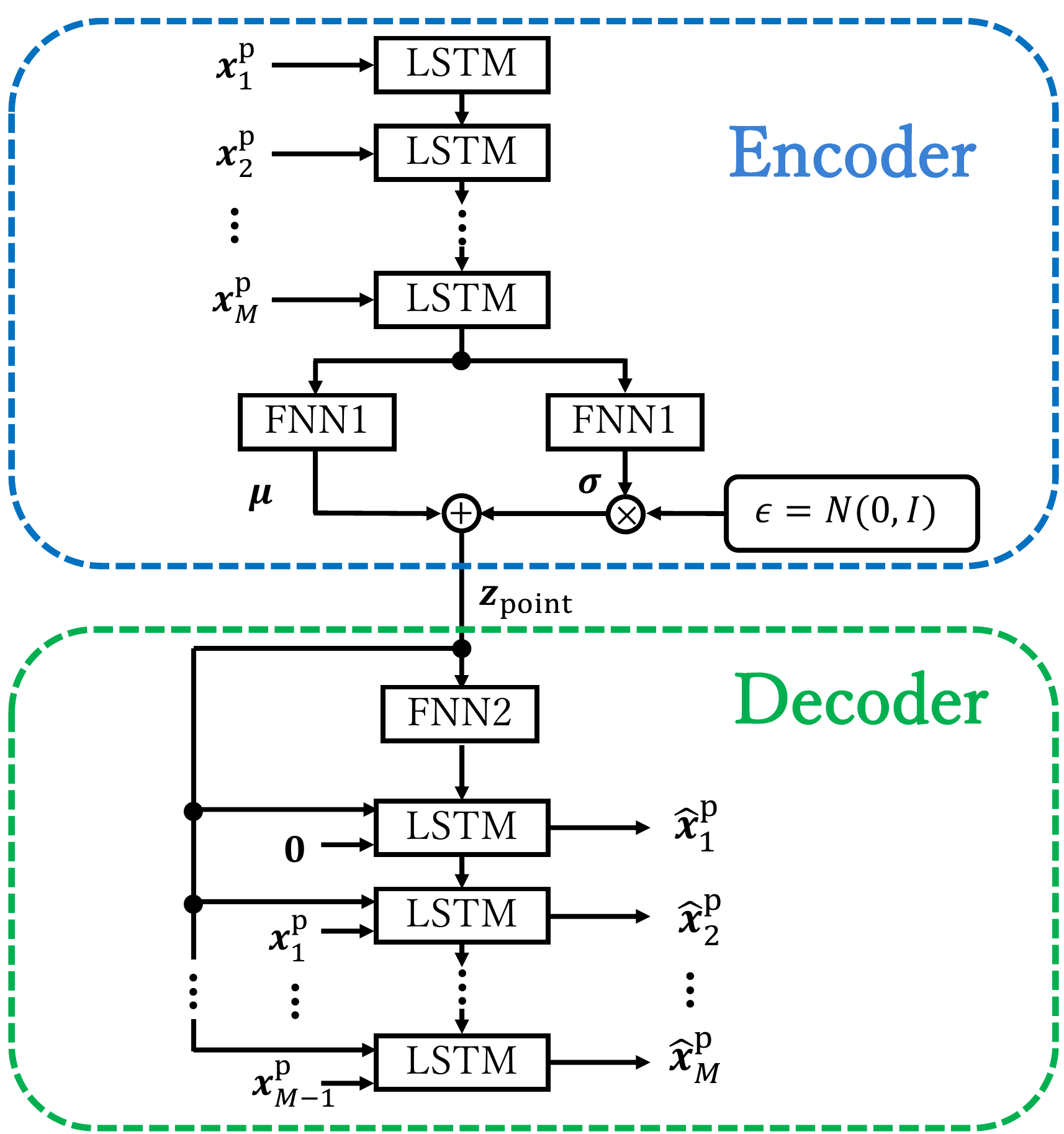}
        \caption{\leftline{ Construction of $\mathrm{VAE_{point}}$.}}
        \label{fig:point}
\end{figure}

\begin{figure}[t]
    \centering
        \includegraphics[width=50mm]{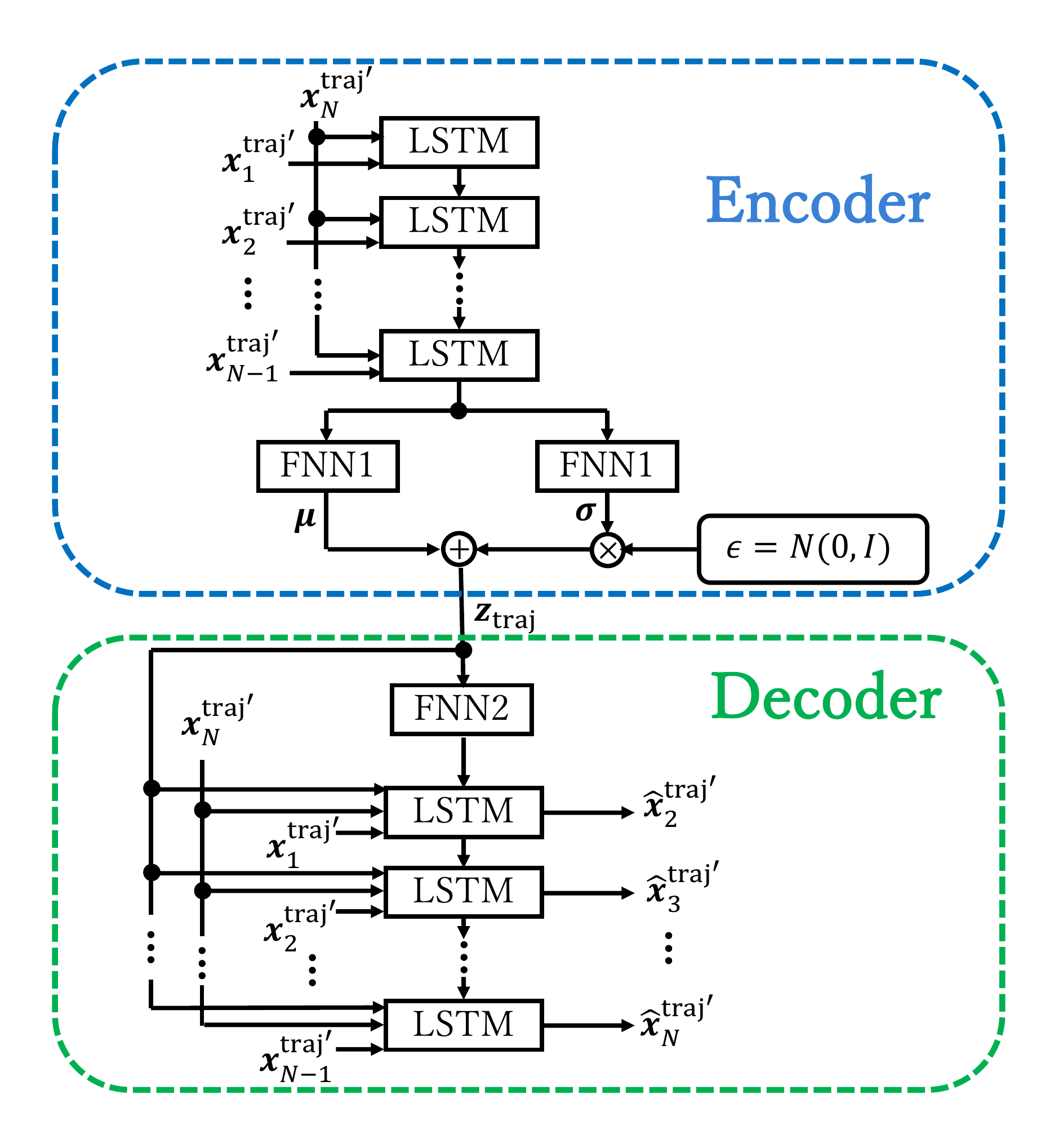}
        \caption{\leftline{ Construction of $\mathrm{CVAE_{traj}}$.}}
        \label{fig:traj}
\end{figure}

In this study, the model that learns the stroke order and generates the start and end points of the stroke trajectory is denoted as $\mathrm{VAE_{point}}$, and the model that learns the touch during the stroke and generates the trajectory is denoted as $\mathrm{CVAE_{traj}}$.
The motion acquired by $\mathrm{VAE_{point}}$ has been denoted as $\bm{x^{\mathrm{point}}_m}$ and the start and end points extracted from the acquired motion as $\bm{x^{\mathrm{start}}_m}$ and $\bm{x^{\mathrm{end}}_m}$, respectively. The subscript $\bm{m}$ denotes the number of stroke counts. The start and end points is shown together as $\bm{x^{\mathrm{p}}_m}$ and has been expressed through~(\ref{eq:point}).
\begin{equation}
    \label{eq:point}
    \bm{x^{\mathrm{p}}_m} = \Bigl(\bm{x^{\mathrm{start}}_m}, \bm{x^{\mathrm{end}}_m} \Bigr)
\end{equation}
The structure of $\mathrm{VAE_{point}}$ is shown in Fig.~\ref{fig:point}. $\mathrm{VAE_{point}}$ learns the pairs of starting and ending points, that is, the stroke order, as the time series data.  Here, $M$ is the number of stroke trajectories; the loss function $L_{\mathrm{point}}$ for learning the $\mathrm{VAE_{point}}$ has been expressed as~(\ref{eq:loss_point}).
\begin{equation}
    \label{eq:loss_point}
    L_{\mathrm{point}} = \frac{1}{M} \sum_{m=1}^{M} \Bigl(\bm{\hat{x}^{\mathrm{p}}_m} - \bm{x^{\mathrm{p}}_m} \Bigr)^2 - L_{\mathrm{KL}}
\end{equation}
Here, $\bm{\hat{x}^{\mathrm{p}}_m}$ is the output of the decoder.
The first term on the right-hand side is the mean squared error for the input and output of the VAE. $L_{\mathrm{KL}}$ is called the Kullback–Leibler (${\mathrm{KL}}$) divergence, which is a measure of the similarity between the two probability distributions. When expressing the similarity between a latent variable and a normal distribution, if the dimension of the latent variable is $J$, the ${\mathrm{KL}}$ divergence can be expressed as~(\ref{eq:loss_kld}), where $\sigma$ and $\mu$ are the encoder outputs.
\begin{equation}
    \label{eq:loss_kld}
    L_{\mathrm{KL}} = \frac{1}{2} \sum_{j=1}^{J} \Bigl(1 + \mathrm{ln}(\sigma^2) - \mu^2 - \sigma^2 \Bigr)
\end{equation}
The behavior obtained in $\mathrm{CVAE_{traj}}$ has been denoted as $\bm{x^{\mathrm{traj}}_n}$. The subscript $\bm{n}$ denotes the number of samples. Here, the entire $\bm{x^{\mathrm{traj}}_n}$ is offset at the starting point of the trajectory, and $\bm{x^{\mathrm{traj'}}_n}$ is the trajectory after processing, expressed through~(\ref{eq:traj'}).
\begin{equation}
    \label{eq:traj'}
    \bm{x^{\mathrm{traj'}}_n} = \bm{x^{\mathrm{traj}}_n} - \bm{x^{\mathrm{traj}}}_1
\end{equation}
The structure of the $\mathrm{CVAE_{traj}}$ is shown in Fig.~\ref{fig:traj}. $\mathrm{CVAE_{traj}}$ generates a trajectory connecting the two points based on the model after the start and end points are given. If the model is trained with a linear trajectory, it generates a linear trajectory, and if the model is trained with a trajectory of repeating short touches, it generates a short-touch trajectory. In other words, the model learns the part corresponding to the touch. To condition the learning process, $\bm{x^{\mathrm{traj'}}}_N$ is always added to the encoder and decoder. $N$ is the number of samples in the trajectory. The loss function $L_{\mathrm{traj}}$ for learning $\mathrm{CVAE_{traj}}$ is given by~(\ref{eq:loss_traj}), where the first term on the right-hand side is the mean squared error for the input and output of the CVAE, the second term is the mean squared error for the derivative of the input and output of the CVAE, and $L_{\mathrm{KL}}$ is the ${\mathrm{KL}}$ divergence.
\begin{eqnarray}
    \label{eq:loss_traj}
    L_{\mathrm{traj}} &=& \frac{1}{N} \sum_{n=1}^{N} \Bigl(\bm{\hat{x}^{\mathrm{traj'}}_n} - \bm{x^{\mathrm{traj'}}_n} \Bigr)^2 \nonumber\\
    &+& \frac{1}{N} \sum_{n=1}^{N} \Bigl( \bm{\hat{x'}^{\mathrm{traj'}}_n}  - \bm{{x'}^{\mathrm{traj'}}_n} \Bigr)^2 + L_{\mathrm{KL}}
\end{eqnarray}

\subsection{Trajectory Generation using Hierarchical CVAE}
\begin{figure}[t]
    \centering
        \includegraphics[width=80mm]{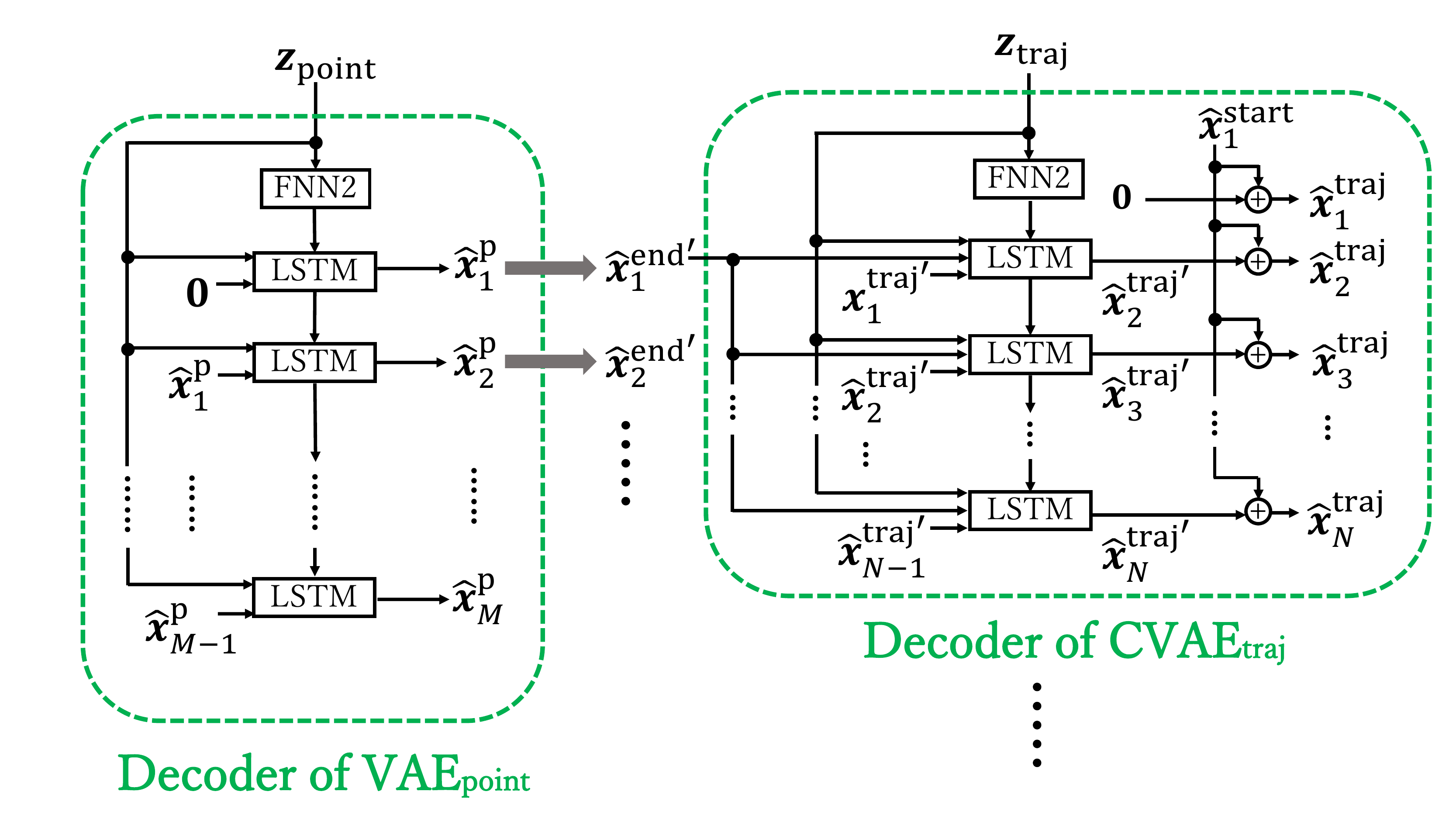}
        \caption{\leftline{ Generation phase.}}
        \label{fig:generate}
\end{figure}

In VAE, trajectories can be generated by separating the encoder and decoder after training the model and assigning latent variables to the decoder. Therefore, after the training of $\mathrm{VAE_{point}}$ and $\mathrm{CVAE_{traj}}$ is completed, the trajectory is generated by combining the decoders of the two models. The structure during the generation is shown in Fig.~\ref{fig:generate}.

First, the latent variable $\bm{z_{\mathrm{point}}}$ is given to the decoder of the $\mathrm{VAE_{point}}$. The start point $\bm{\hat{x}^{\mathrm{start}}}$ and end point $\bm{\hat{x}^{\mathrm{end}}}$ of the trajectory are the output, the endpoint $\bm{\hat{x}^{\mathrm{end'}}}=\bm{\hat{x}^{\mathrm{end}}}-\bm{\hat{x}^{\mathrm{start}}}$ is obtained by offsetting the starting point with the output start point. We mapped this value to the endpoint of the $\mathrm{CVAE_{traj}}$. The obtained $\bm{\hat{x}^{\mathrm{end'}}}$ and the latent variable $\bm{z_{\mathrm{traj}}}$ are the input to the decoder of $\mathrm{CVAE_{traj}}$, and the starting point $\bm{\hat{x}^{\mathrm{start}}}$, which is subtracted at the offset, is added to the output trajectory to generate a trajectory that combines the two models.

\subsection{Data padding process}
In this study, we recorded and learned human drawing and grinding motions; however, unlike images and sounds, the amount of training data is greatly limited for practical use. However, generalization performance to cope with various conditions is often required in actual robots. For example, the same character is drawn at different positions requires various levels of processing in a robot program, which is time consuming. 

To be able to generate trajectories under various conditions for different positions and angles of the drawn lines, it is necessary to obtain multiple sets of data so that the generalization performance can be obtained for each assumed condition. As the amount of these conditions increases, the number of training data required increases exponentially, making robot learning difficult in practice. Data with these different conditions can be obtained through simple geometric transformations such as translation and rotation, in this study, we train the robot by padding the human training data with geometric transformations. One of the disadvantages of this method is that the appropriate variance of the brush strokes cannot be obtained owing to the small amount of data. In this study, 10 human motions were obtained. This was the minimum number and did not pose a risk.

In addition, the length of the line drawn by a single stroke cannot be simply increased because the size and frequency of the brush strokes change when the training data are generated in proportion to the length of the line through geometric transformation.
Therefore, in this study, we trained long and short lines respectively to obtain the generalization performance against the length variation.

\begin{figure}[t]
        \centering
        \includegraphics[width=80mm]{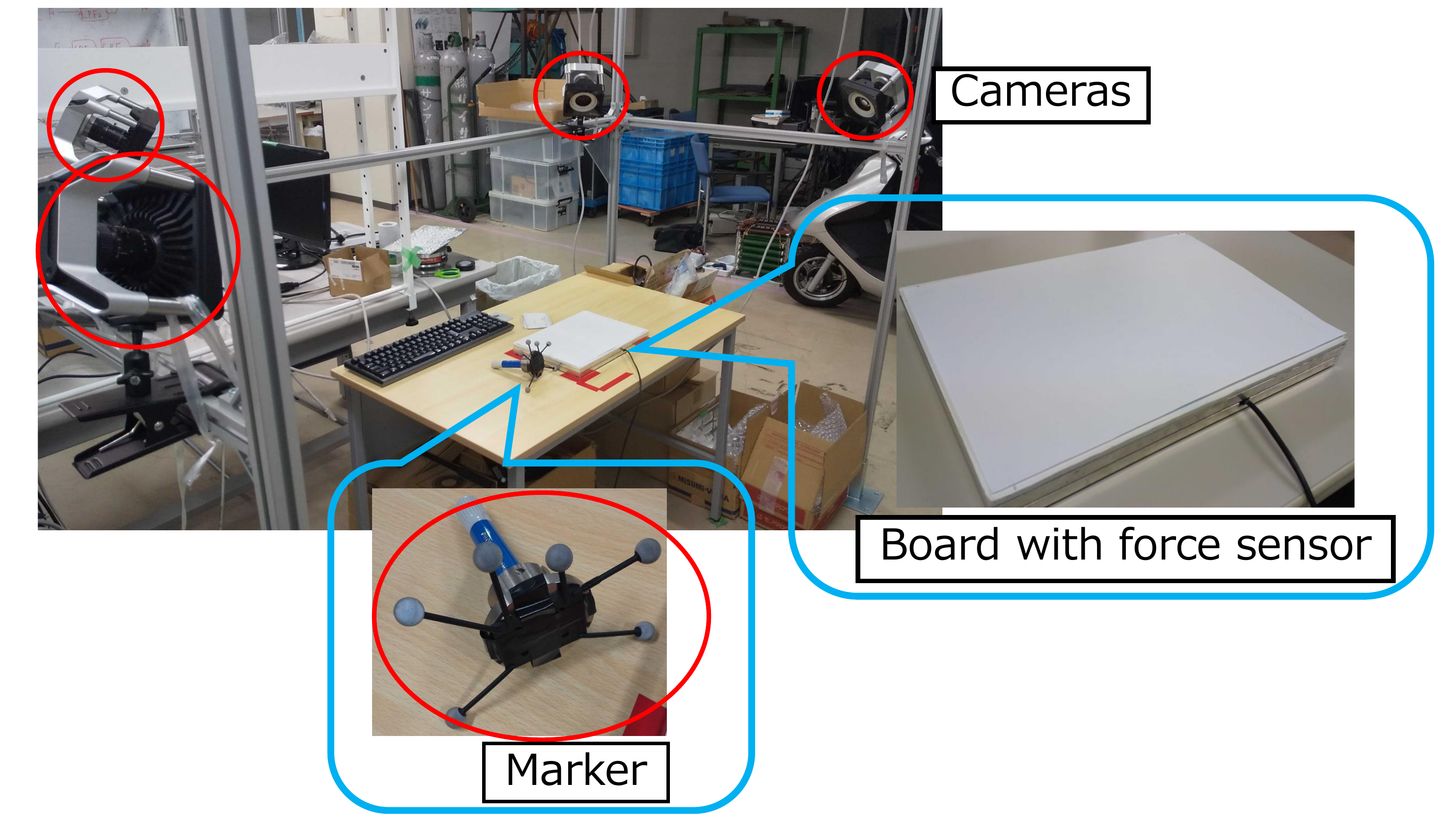}
        \caption{\leftline{ Experimental setup for acquiring human motion.}}
        \label{fig:optitrack}   
\end{figure}
    
\section{Experiment}
In this section, we describe the experiments conducted to verify the effectiveness of the proposed method.
\subsection{Collection of Teaching Data}\label{Experiment}
In this study, we used the OptiTrack~\cite{optitrack} and a foundation-type sensor embedded with a six-axis force sensor~\cite{totsu2016} to record the human motion. The recording period of the OptiTrack and force sensor was 8~ms. Fig.~\ref{fig:optitrack} shows the device used. When a human performs an action using a tool with a marker in Fig.~\ref{fig:optitrack}, the position and force information during the action are recorded. By acquiring actions that use the same tools performed by humans, we can record pure human actions without including the constraints of a robot. 
    
\begin{figure}[t]
    \begin{minipage}[b]{0.48\linewidth}
        \centering
            \includegraphics[width=43mm]{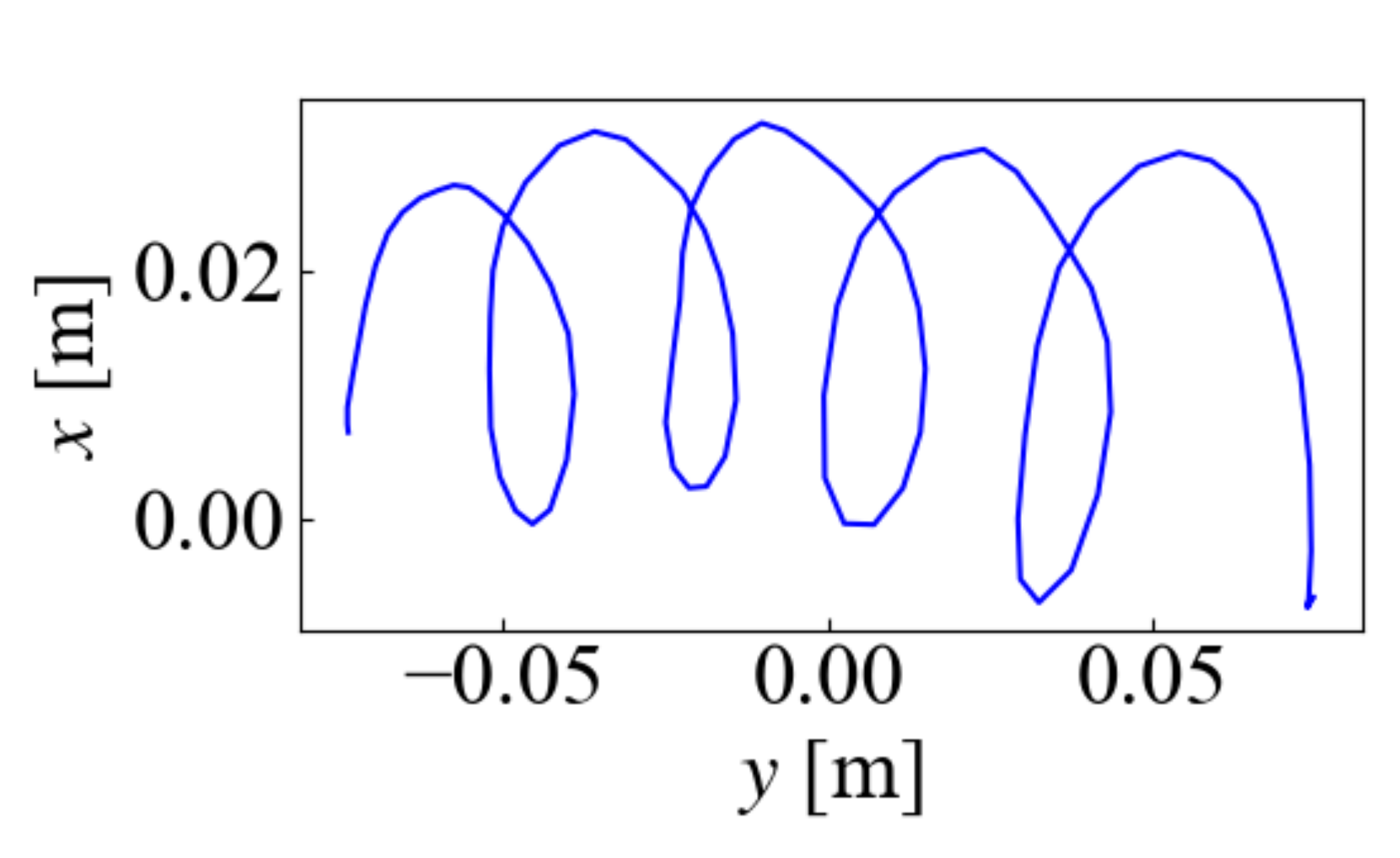}
            \subcaption{Long trajectory}%
            \label{fig:long}       
    \end{minipage}
    \begin{minipage}[b]{0.48\linewidth}
        \centering
            \includegraphics[width=43mm]{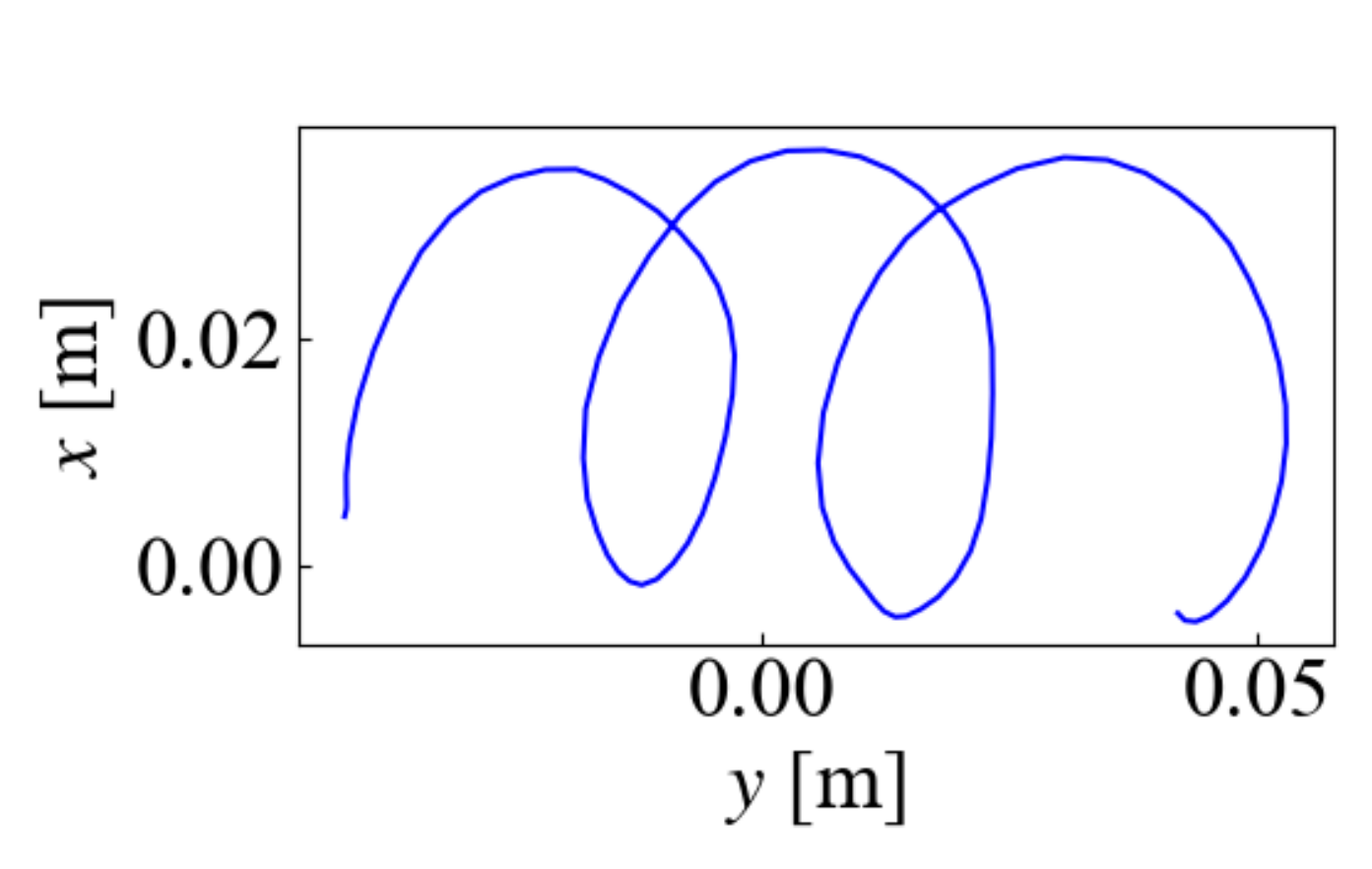}
            \subcaption{Short trajectory}
            \label{fig:short}        
    \end{minipage}
    \caption{\leftline{ Trajectories acquired by human.}}
    \label{fig:guruguru}
\end{figure}

For the motion data generated using a pen, we obtained trajectory \textcircled{\scriptsize 1} $\mathrm{"A"}$, 20 times. Trajectory \textcircled{\scriptsize 2} shown in Fig.~\ref{fig:long} and the shorter trajectory shown in Fig.~\ref{fig:short} were obtained 10 times each. In the data acquired in \textcircled{\scriptsize 1}, (a) "the action of the tool touching and drawing a line" and (b) "the movement from the end of the line to the start of the next line" were included. Therefore, to obtain the stroke order of characters, the range of (a) was defined as the area where the contact force was 0.25 N or higher; the range of (b) was eliminated, and the starting and ending points of the characters were extracted. In the case of $\mathrm{"A"}$, the number of trajectories is three because it consists of three lines, $M = 3, m = 1, 2, 3$. In addition, the learning cost may have been an issue had we used the data obtained in \textcircled{\scriptsize 2}. Therefore, the acquired data was down-sampled to obtain 100 samples. Thus, $N = 100, n = 1, 2, \cdots, 100 $. Furthermore, we would likely not learn well with data in a single direction only. Therefore, we rotated the 10 data by $20^\circ$and increased the total number of data to 180.
    
\begin{table}[t]
    \renewcommand{\arraystretch}{1.3}
    \caption{PARAMETERS OF VAES}
    \label{table:vae_parameter}
    \centering
      \begin{tabular}{|c|c|c|}
        \hline
         & $\mathrm{VAE_{point}}$ & $\mathrm{CVAE_{traj}}$ \\
        \hline
        LSTM layer size & 1 & 2 \\
        \hline
        LSTM unit & 64 & 256 \\
        \hline
        Dimension of $z_{\mathrm{point}}$/$z_{\mathrm{traj}}$ & 6 & 3 \\
        \hline
        FNN2 activation function & tanh & tanh \\
        \hline
        Optimizer & Adam & Adam \\
        \hline
        Batch size & 4 & 10 \\
        \hline
        Epoch  & 10000 & 50000 \\
        \hline
       \end{tabular}
\end{table}

\begin{figure}[t]
     \centering
     \includegraphics[width=80mm]{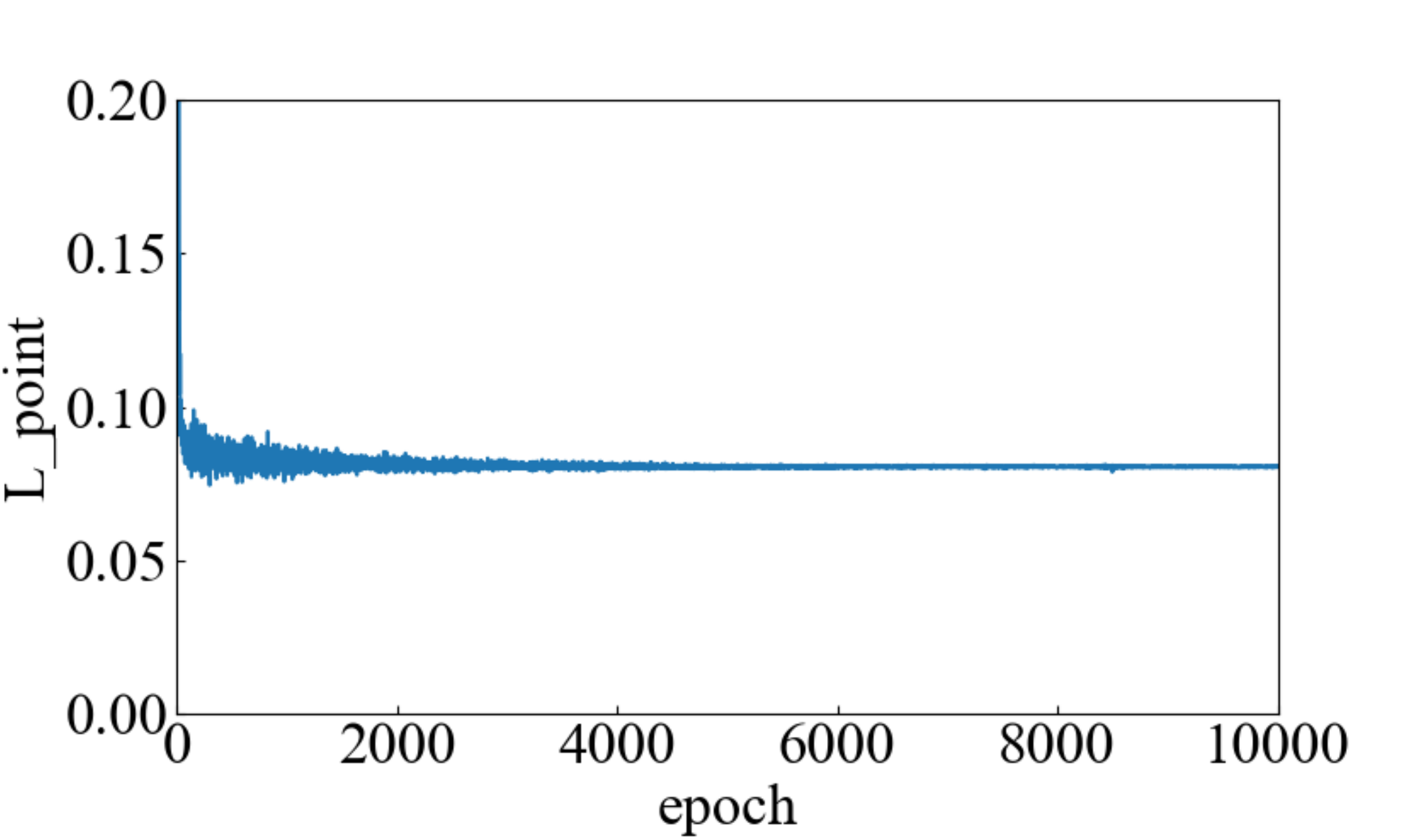}
     \caption{\leftline{ Loss of $\mathrm{VAE_{point}}$.}}
     \label{fig:loss_A}
\end{figure}
\begin{figure}[t]
     \centering
     \includegraphics[width=80mm]{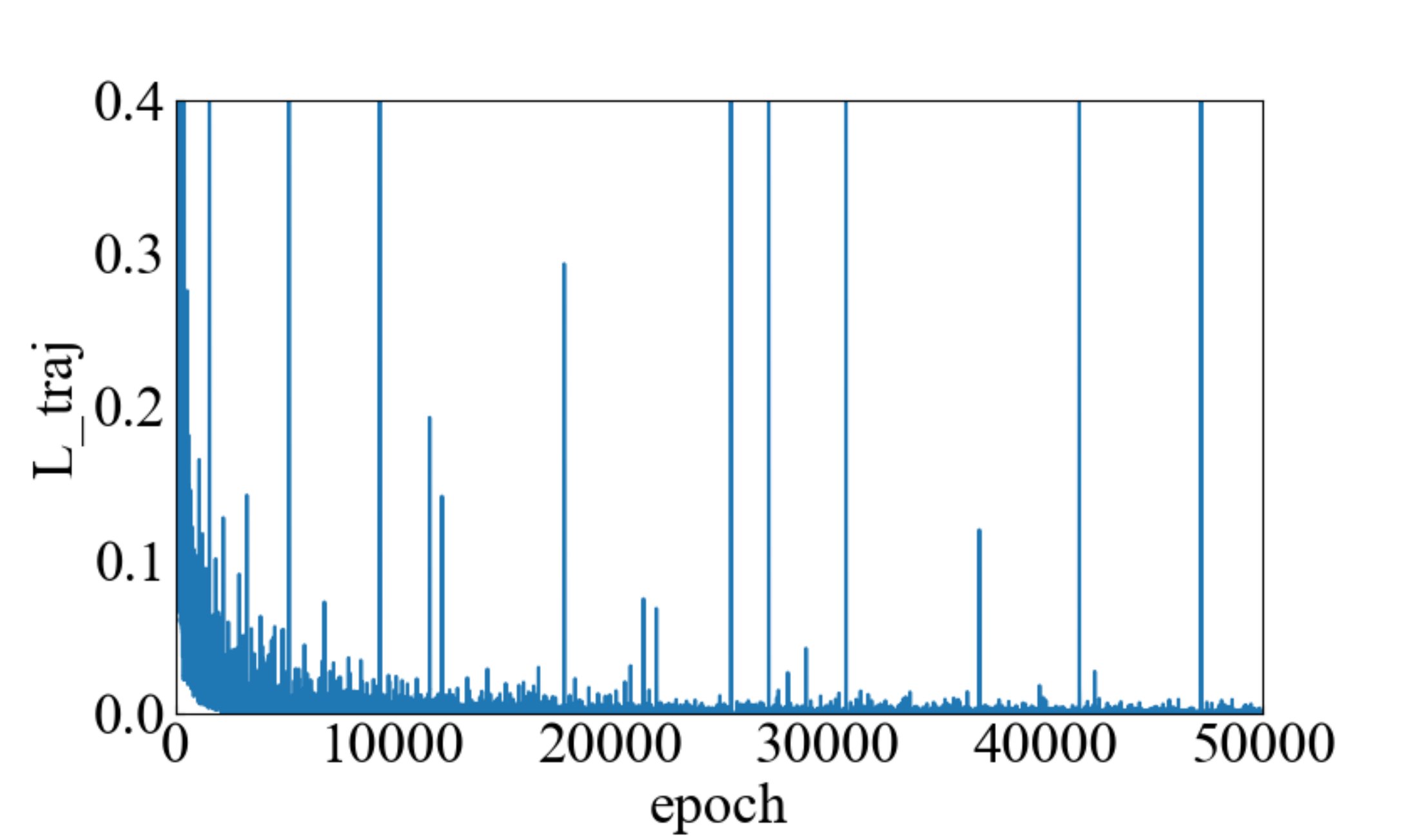}
     \caption{\leftline{ Loss of $\mathrm{CVAE_{traj}}$.}}  
   \label{fig:loss_traj}
\end{figure}

\begin{figure}[t]
  \centering
      \includegraphics[width=85mm]{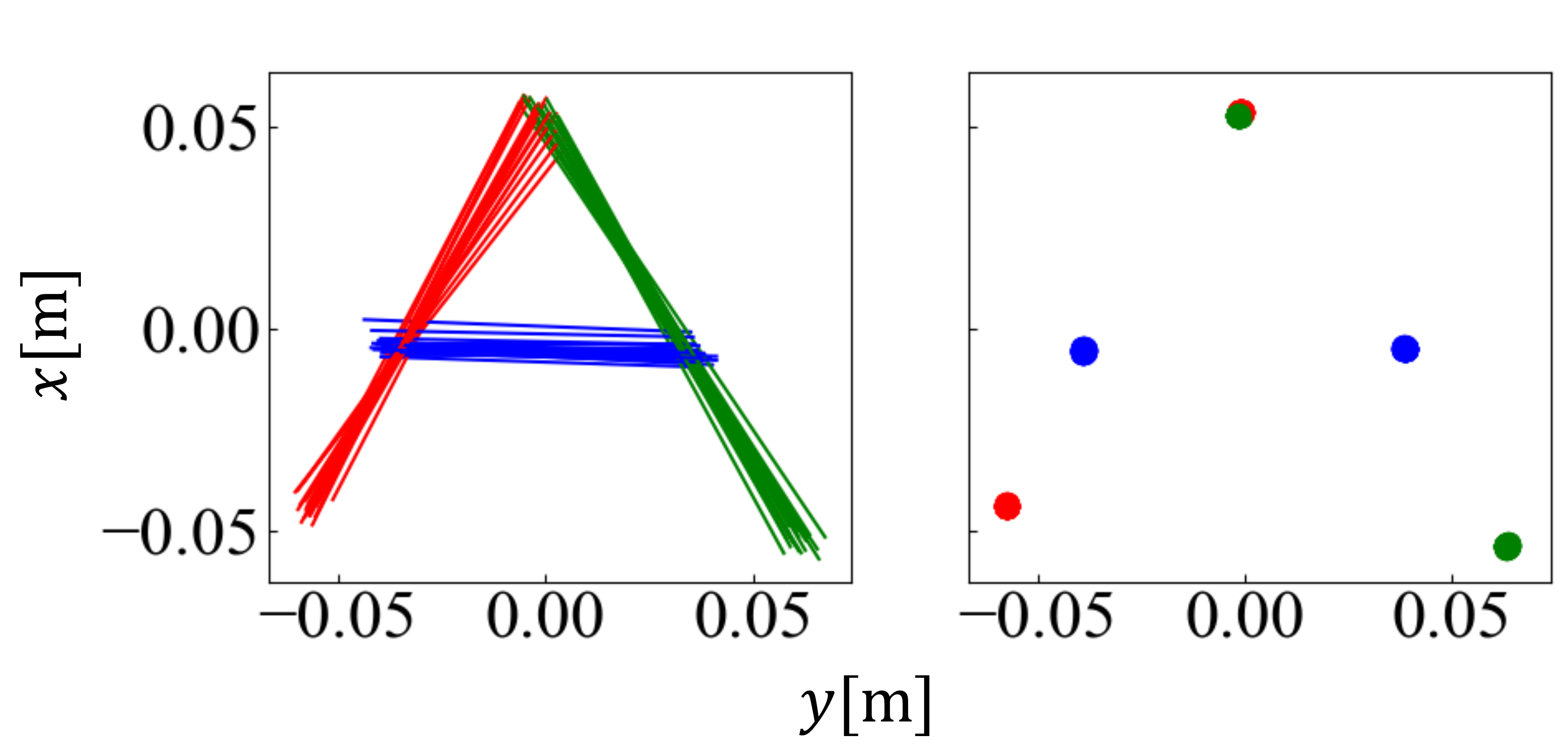}
      \caption{\leftline{ $\mathrm{VAE_{point}}$ input/output.}}
      \label{fig:train_A}
\end{figure}
    
\begin{figure}[t]
    \centering
        \includegraphics[width=85mm]{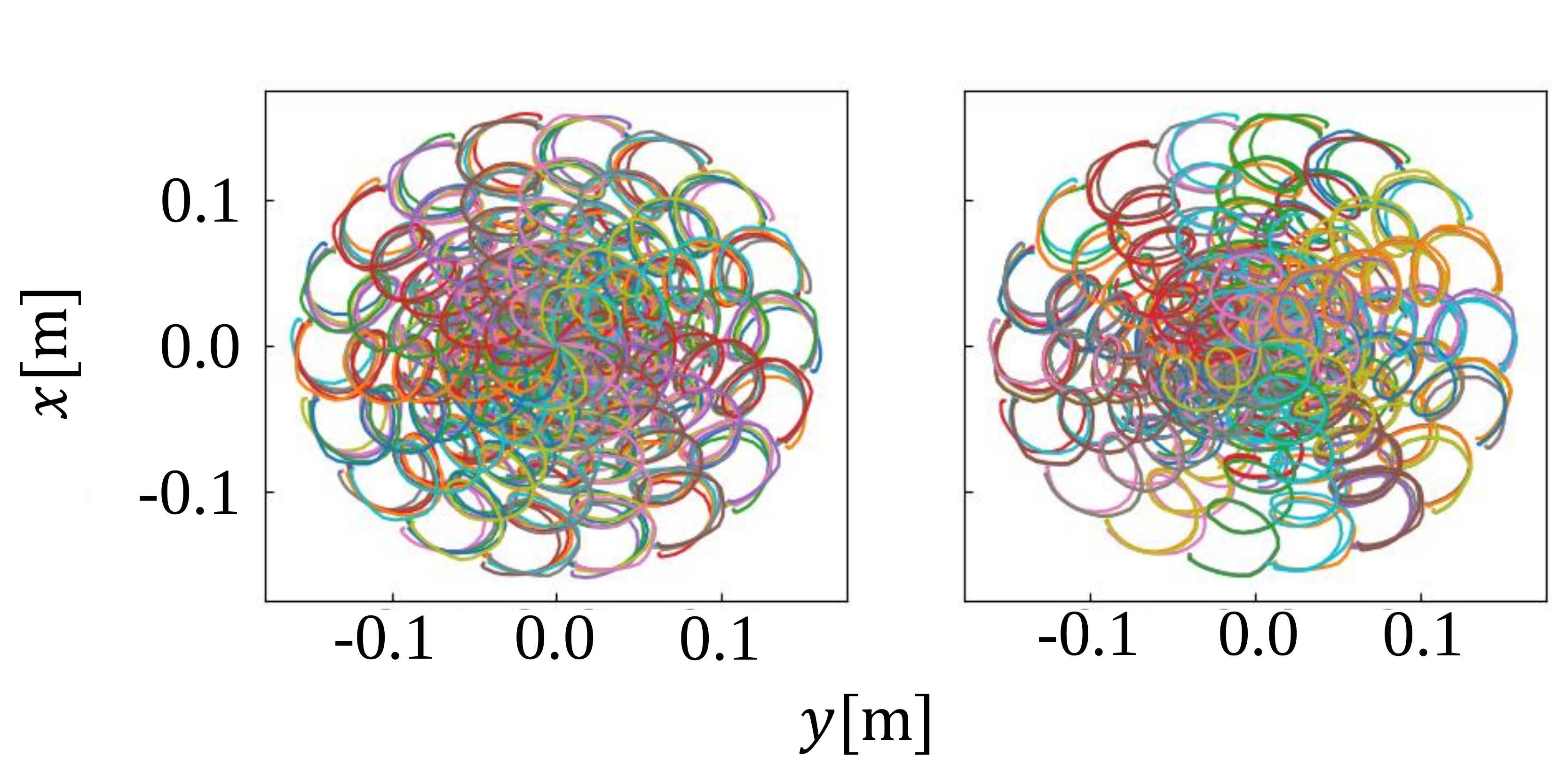}
        \caption{\leftline{ $\mathrm{CVAE_{traj}}$ input/output.}}
        \label{fig:train_traj}
\end{figure}
    
\begin{figure}[t]
  \centering
      \includegraphics[width=80mm]{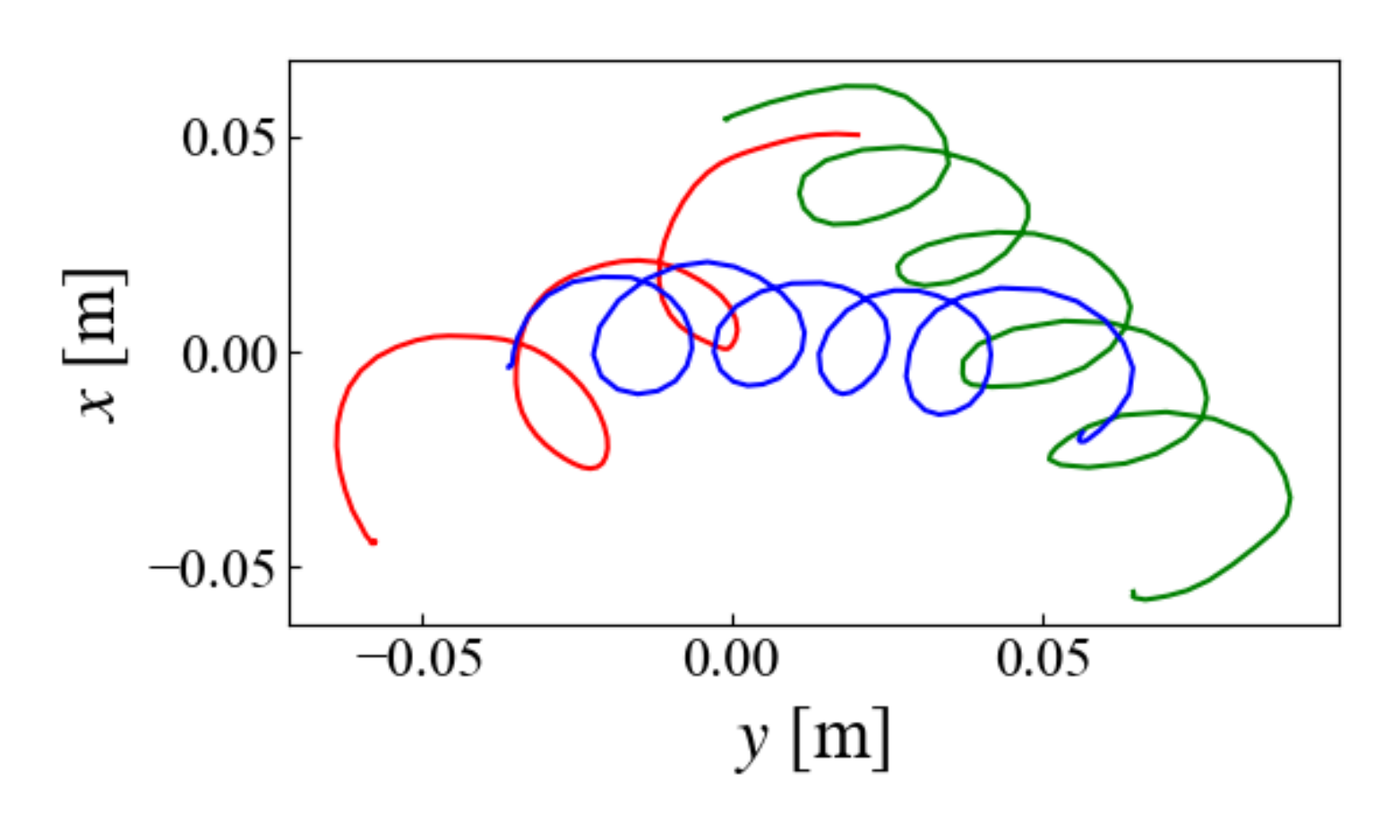}
      \caption{\leftline{ Output result of the combination.}}
      \label{fig:result_A}
\end{figure}
\begin{figure}[t]           
  \centering
    \includegraphics[width=90mm]{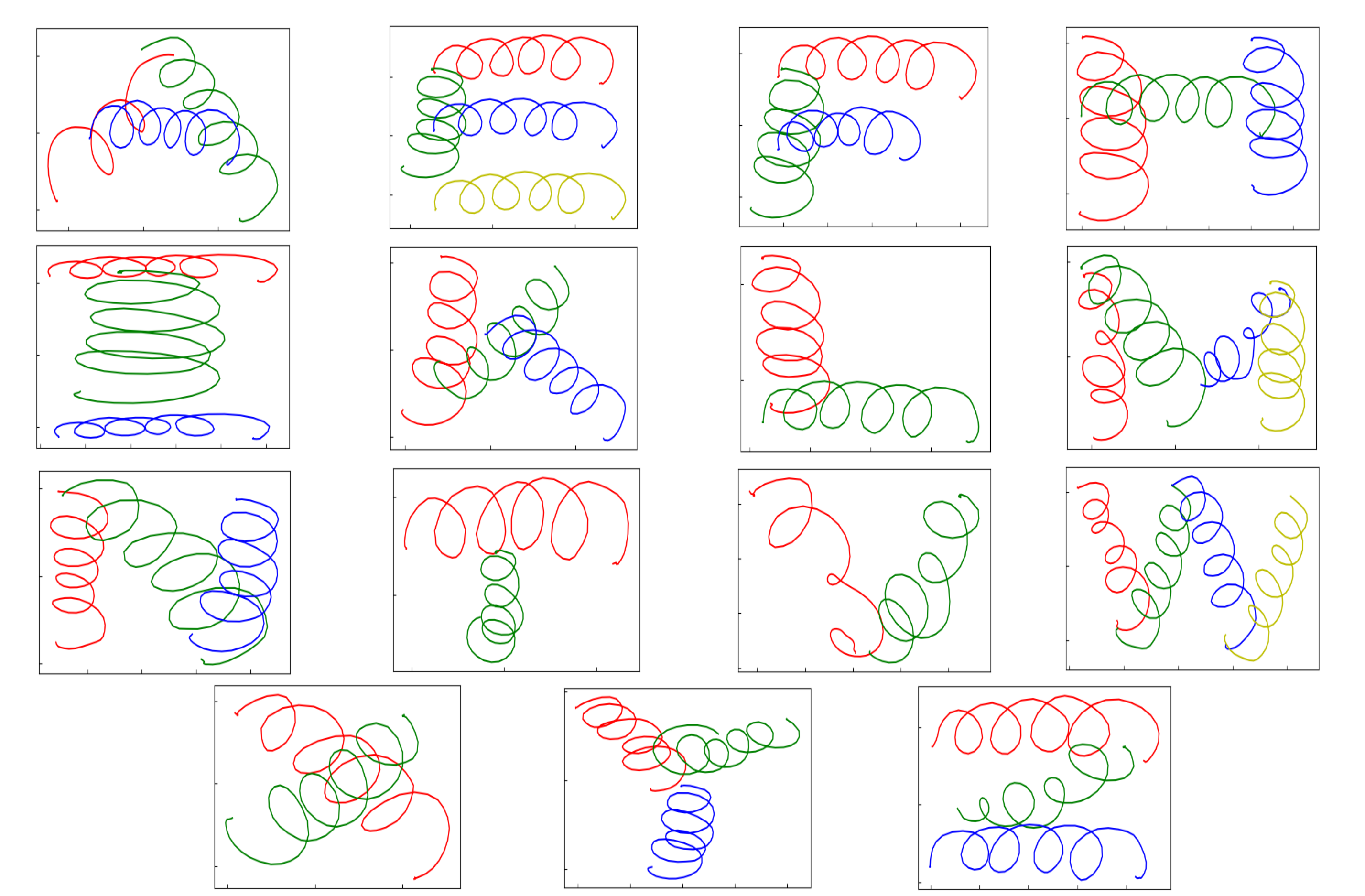} 
  \caption{ Results of using 15 different characters\\(A,E,F,H,I,K,L,M,N,T,V,W,X,Y,Z).} 
  \label{fig:generatechara}
\end{figure}
    
\subsection{Experimental Results}\label{combi}
The parameters of the learning models used in this study are shown in Table~\ref{table:vae_parameter}. Figs.~\ref{fig:loss_A} and~\ref{fig:loss_traj} show the change in loss during the training of the two models. Figs.~\ref{fig:train_A} and~\ref{fig:train_traj} show the results when the two models were generated independently of each other. 
In these figures, the left side is the training data and the right side is the output. The input and output in the $\mathrm{VAE_{point}}$ are the start and end points, respectively, and these two points are shown in Fig.~\ref{fig:train_A}. Noticeably, the three sets of start and end points are the output according to the stroke order. 
In addition, $\mathrm{CVAE_{traj}}$ shows that long and short trajectories are the output. 
Fig. 11 shows the results generated by combining the two models that were trained independently. Notably, the brush strokes learned by $\mathrm{CVAE_{traj}}$ corresponded to the start and end points output by $\mathrm{VAE_{point}}$.
    
\subsection{Changing the input to $\mathrm{VAE_{point}}$}
We prepared the trajectory information of 15 characters as input data for $\mathrm{VAE_{point}}$ and the trajectories were generated without changing the $\mathrm{CVAE_{traj}}$ model. The results of the combination of the two models are shown in Fig.~\ref{fig:generatechara}. It can be seen that the trajectories generated by the $\mathrm{CVAE_{traj}}$ model correspond to the starting and ending points of the $\mathrm{VAE_{point}}$ model, even for different letters, such as $\mathrm{"A"}$. Moreover, the brush strokes learned by $\mathrm{CVAE_{traj}}$ did not learn the vertical trajectory, In Fig.~\ref{fig:generatechara}, the vertical trajectory is also reproduced by the brush strokes corresponding to the start and end points output by $\mathrm{VAE_{point}}$ as well as the other trajectories.

\subsection{Changeing input to $\mathrm{CVAE_{traj}}$}\label{change_traj}
\begin{figure}[t]           
  \centering
    \includegraphics[width=80mm]{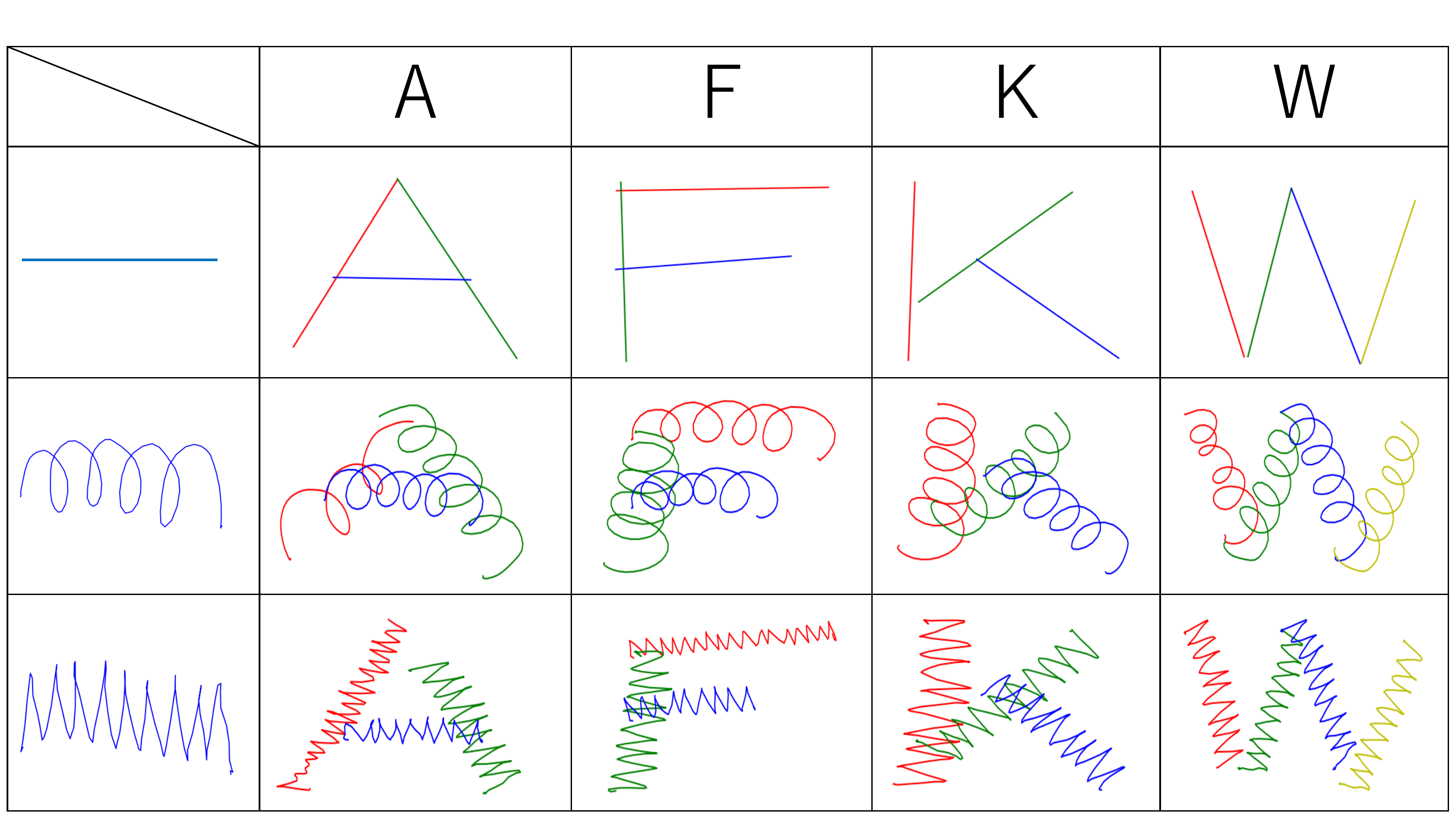}
  \caption{\leftline{ Output results of replacing the models.}}
  \label{fig:three}
\end{figure}

In this section, we examine what changes appear in the generated trajectories by replacing the models in the lower layers, which learns touch. Fig.~\ref{fig:three} shows the results when several characters are drawn with different models in the lower layer. The first row shows the character used for the upper layer, and the first column shows the touch used in the lower layer. These results show that by using models with different touches, we can easily generate more patterns of output by rearranging the hierarchy.

Next, the similar trajectories to those in Fig.~\ref{fig:guruguru}, which were obtained using the pen in \ref{Experiment}, were obtained using the grinding tool. For the stroke order of the letters, the trajectories were generated by using the learning models of several $\mathrm{VAE_{point}}$ acquired with the pen and combining them with the model of $\mathrm{CVAE_{traj}}$ acquired with the grinding tool. Fig.~\ref{fig:W} shows the trajectories of $\mathrm{"W"}$ generated when both the stroke order and touch are learned with a pen and when the touch was learned with a grinding tool. The output of the generated force is illustrated in Fig.~\ref{fig:compare}. It was found that the output results of the same character changed according to the difference in the learned brush strokes. Therefore, we can increase the number of trajectories generated by using different tools for stroke orders and brush strokes.

\subsection{Experiments with a real robot}
This section describes the generated trajectory replayed on a real robot and grinded on a metal plate to confirm its adaptability to the real environment. The system used in the experiment is shown in Fig.~\ref{fig:robot}. 
A force sensor was mounted on at end of the six-degree-of-freedom (6-DOF)-manipulator; the grinding tool and gripper were replaceable. In this experiment, the force in the $z$-direction was used as the contact force, and an impedance-controlled system is used as the control system. A block diagram of the control system is presented in Fig.~\ref{fig:control}. A disturbance observer (DOB)~\cite{dob} was used to guarantee the disturbance of the control system for position and force. The parameters of the control system are listed in Table~\ref{table:control}.

\begin{figure}[t]
    \begin{minipage}[t]{1.0\linewidth}
      \centering
      \includegraphics[width=70mm]{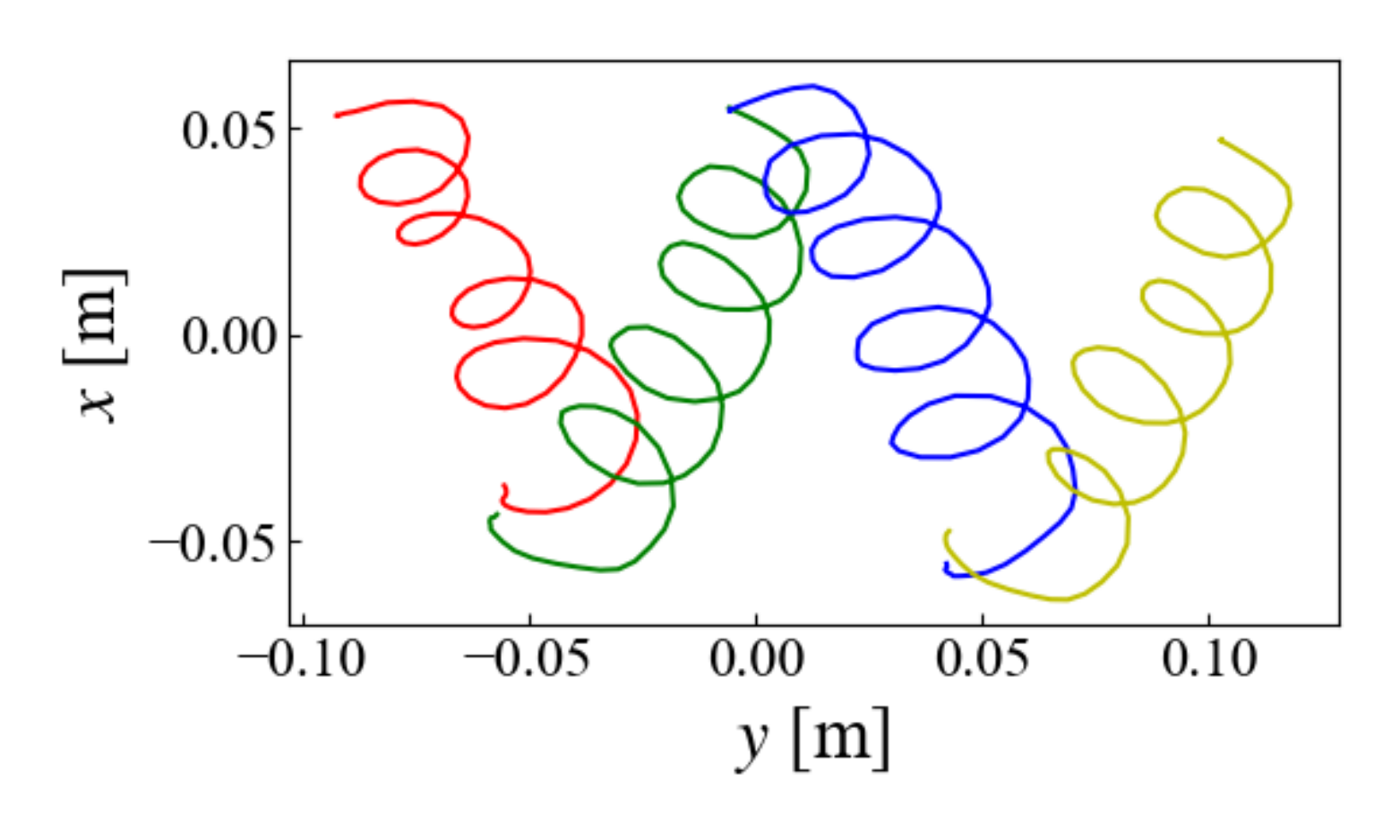}
      \subcaption{Output using pen's data}
      \label{fig:W_pen}
    \end{minipage}
    \begin{minipage}[t]{1.0\linewidth}
      \centering
      \includegraphics[width=70mm]{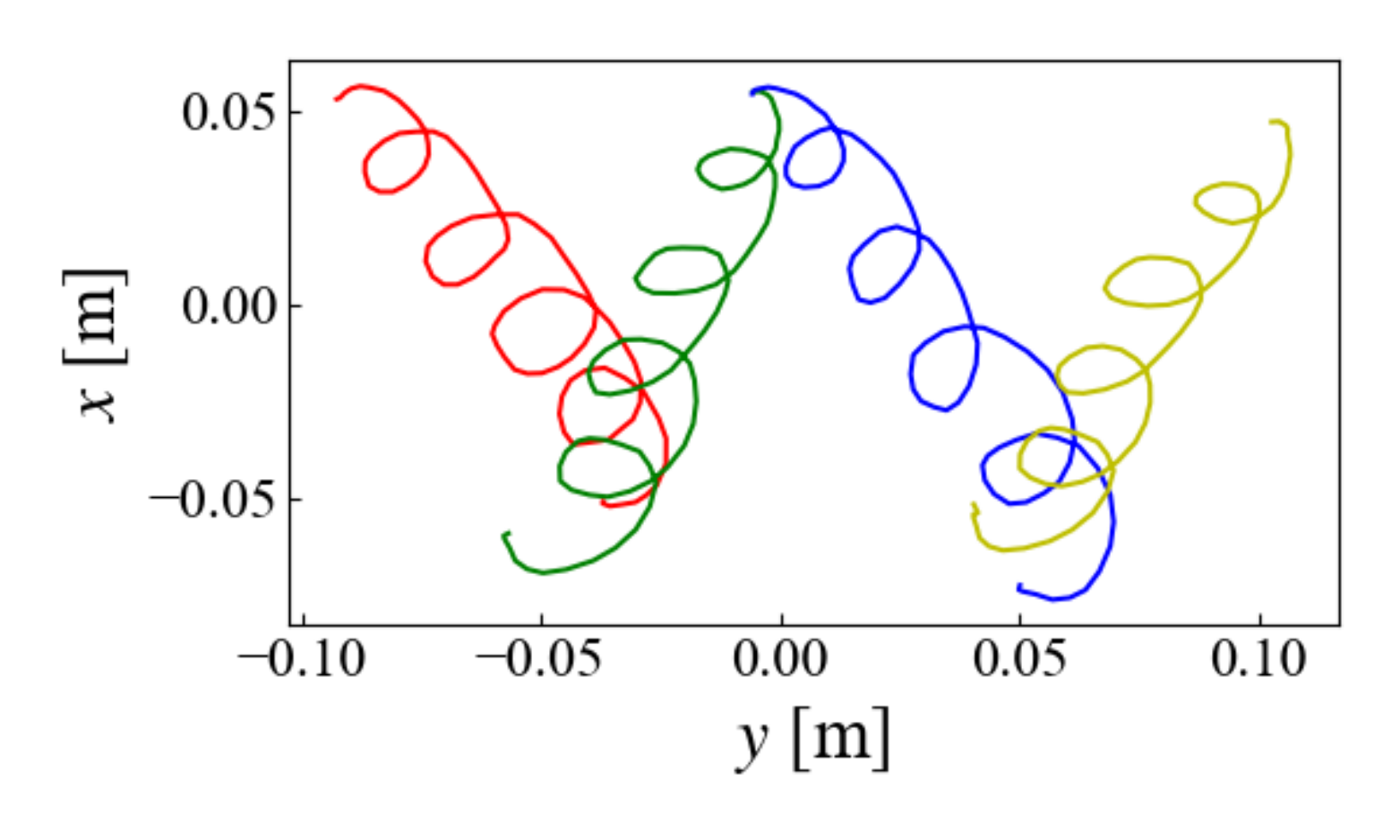}
      \subcaption{Output using grinding data}
      \label{fig:W_polish}
    \end{minipage}
    \caption{\leftline{ Comparison of outputs with different tools.}}
    \label{fig:W}
\end{figure}

\begin{figure}[t]
    \centering
        \includegraphics[width=80mm]{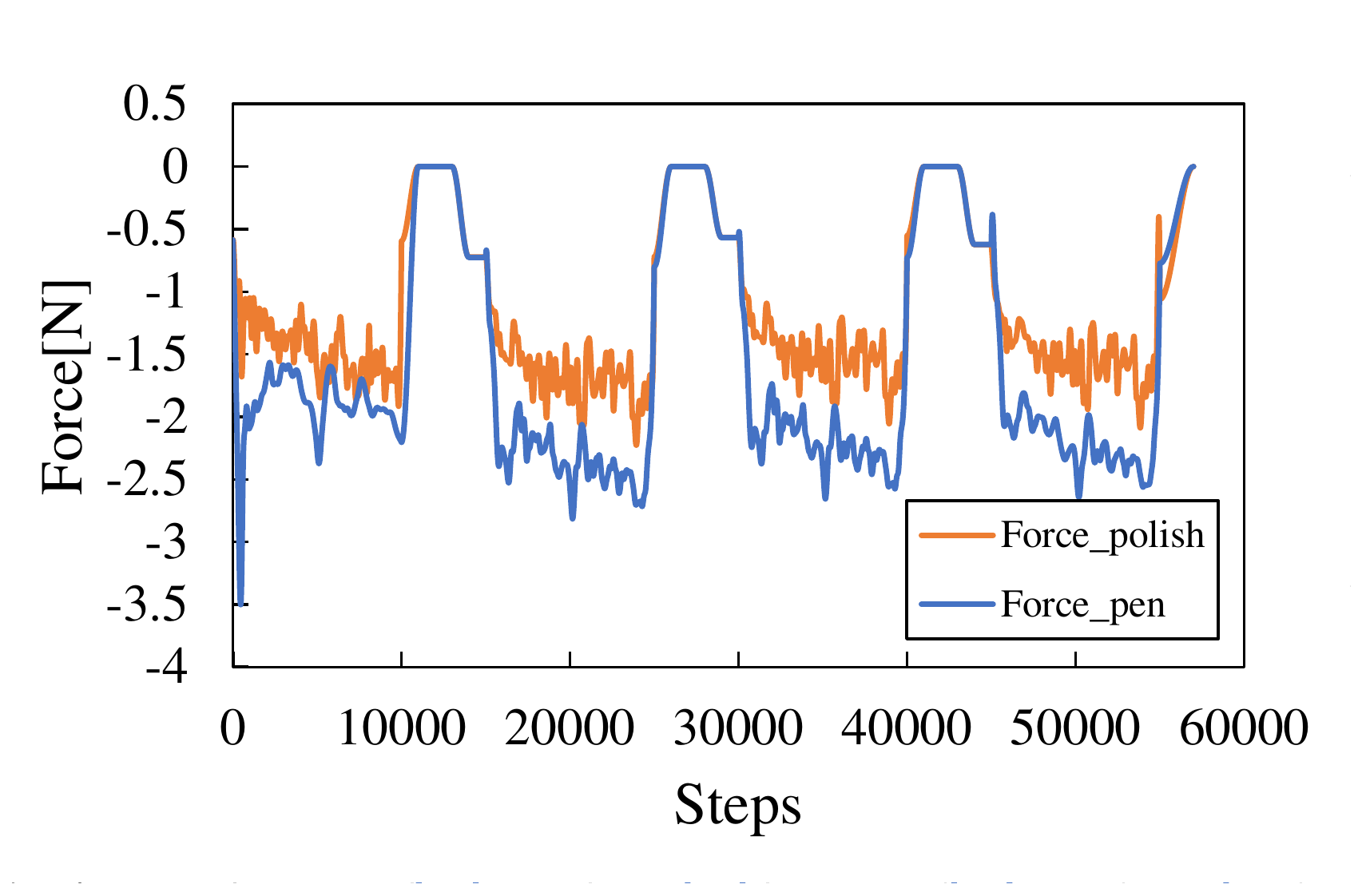}
        \caption{\leftline{ Outputs of force information from different tools.}}
        \label{fig:compare}
\end{figure}

\begin{figure}[t]
    \centering
        \includegraphics[width=80mm]{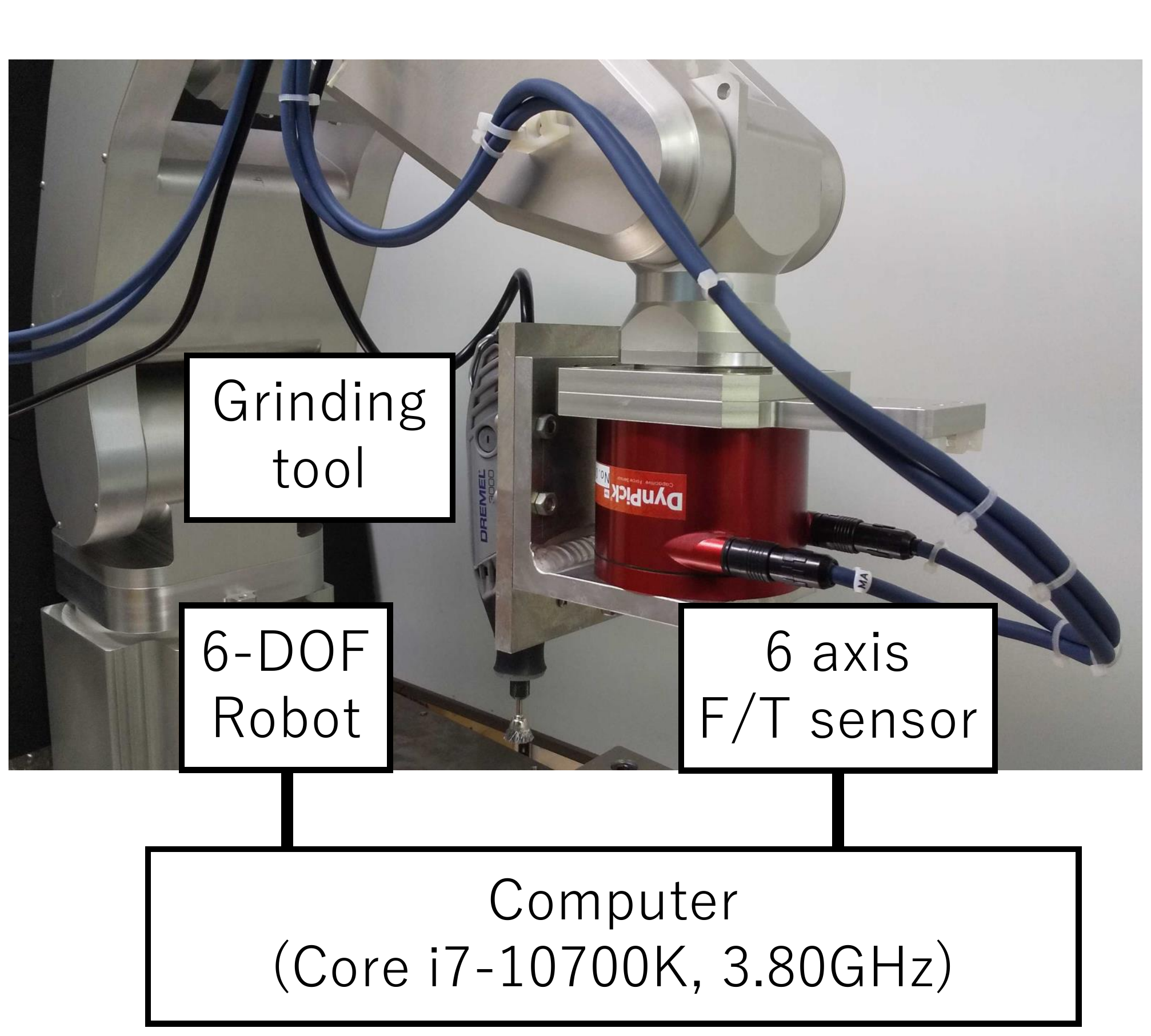}
        \caption{\leftline{ 6-DOF manipulator.}}
        \label{fig:robot}    
\end{figure}

\begin{figure}[t]
  \centering
      \includegraphics[width=80mm]{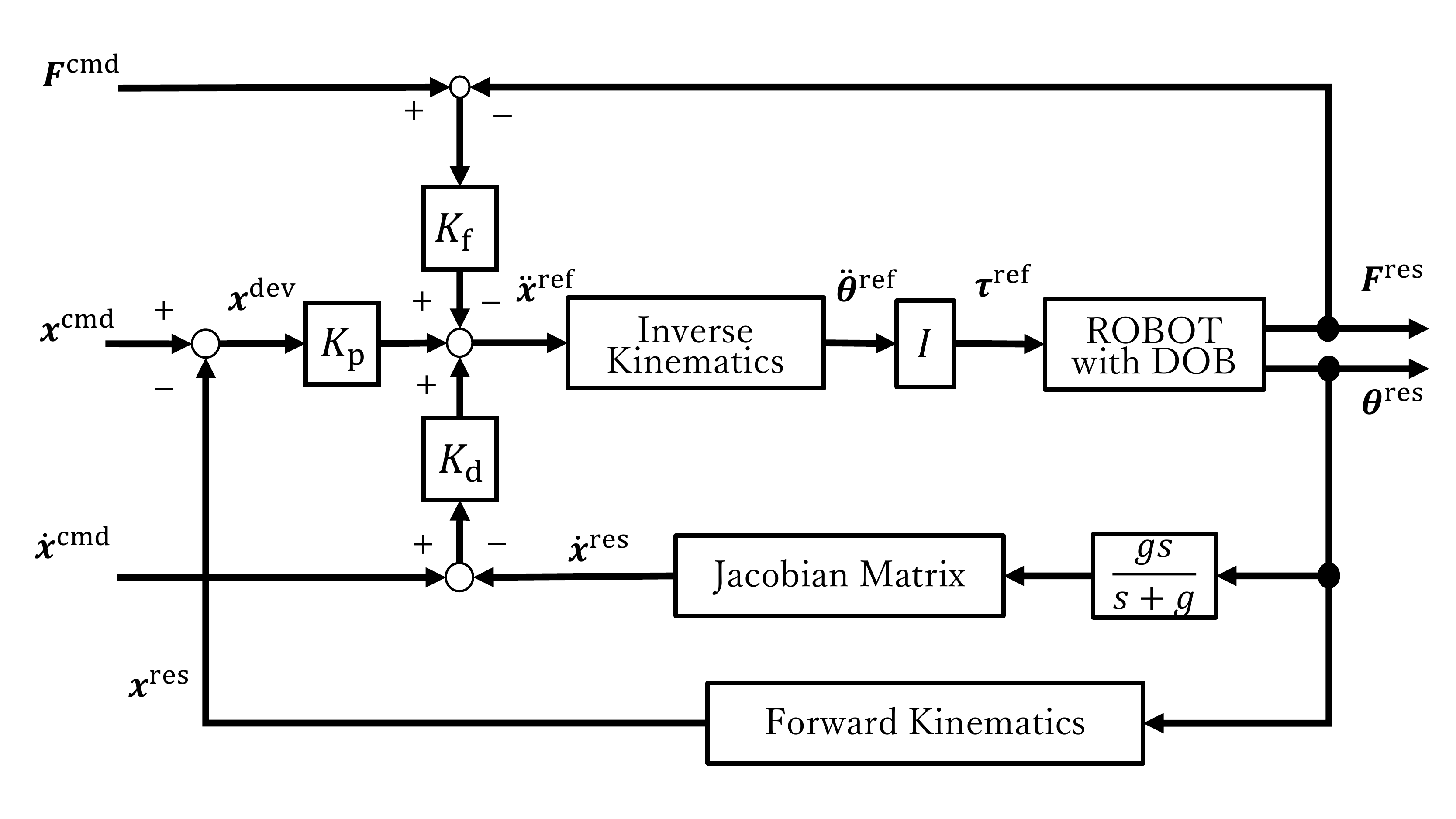}
      \caption{\leftline{ Block diagram of control system.}}
      \label{fig:control}
\end{figure}

\begin{figure}[t]
    \centering
        \includegraphics[width=80mm]{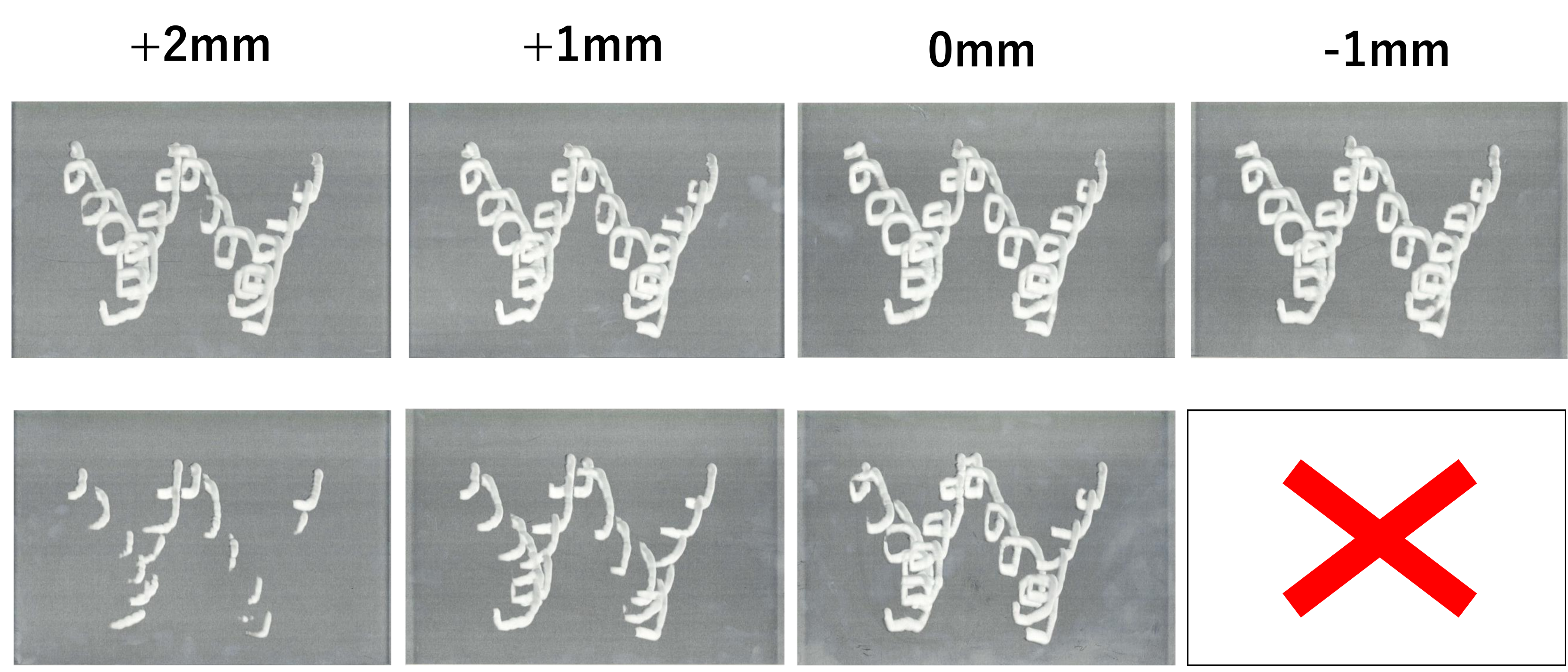}
        \caption{\leftline{ Comparison with and without force control.}}
        \label{fig:robotW}
\end{figure}

The sampling time was set to 1~ms. Since the interval time of the output obtained in Fig.~\ref{fig:W} is longer, we used spline interpolation. The trajectories ($x, y, f_{z}, v_{x}, v_{y}$) obtained here were input to the robot as commands, and the trajectories were replayed using the grinding tool when the robot learned the stroke order and touch by grinding. 
To show the usefulness of the output force information, we compared the results with those obtained by using a control without force information. In addition to the reference height, experiments were conducted when the height was increased by 1~mm and 2~mm, respectively, and when the height was set to -1~mm on the force control side. In the case of control without the learned force information, we did not set the command to -1~mm because it would cause excessive force. The upper panel of Fig.~\ref{fig:robotW} shows the results with force information, and the lower panel shows the results without force information. 
This comparison confirms that the inclusion of force information in the trajectory generation is important to reproduce the output according to the learned differences in touch. 
Because the trajectories were generated by combining two models, it is possible to reproduce trajectories that are not actually drawn by a person on a real robot, and the number of trajectories that can be combined and output can be easily increased by changing the learning data of the stroke order and touch.

\section{Conclusion}
In grinding and drawing, global motion (stroke order) is achieved by repeating the local reciprocating motions (touches) that make up the periodic pattern. 
In this study, we proposed a method to model the stroke order and touch with the VAE. 
By combining these two separately trained, in a hierarchical structure to generate trajectories, it is possible to generate higher-order trajectories with high reproducibility for both local and global features. 
By changing the combination of the learned models, it is possible to generate new trajectories that have never been learned in the past. 

From the simulation and experimental results, it was confirmed that the CVAE model of touch generates trajectories corresponding to the start and end point information given as labels. 
Furthermore, it was shown that the CVAE has the ability to interpolate and generate trajectories with lengths and angles in between, whereas the training data of touch acquired from a person consists of two types of lengths and three types of angles. 

It was shown that a generation network with the same structure could be made into a multistage network, which enables the generation of higher-order trajectories with a relatively small amount of training data. 
In this study, we realized the automatic generation of tasks subject to mechanical constraints from the environment for planar grinding; To generalize this technology, future studies should extend it to 3D shapes and other tasks.

\begin{table}[t]
  \caption{PARAMETERS OF CONTROLLER}\label{table:control}
  \centering
   \begin{tabular}{l||l|l}
    \hline
    $K_{\mathrm{p}}$ & Propotional gain & diag(500, 500, 100, \\
    & & 600, 600, 600)~[1/$\mathrm{s^2}$] \\
    
    $K_{\mathrm{d}}$ & Derivative gain & diag(35, 35, 200, \\
    & & 60, 60, 60)~[1/$\mathrm{s}$]  \\
    
    $K_{\mathrm{f}}$ & Force gain & diag(0, 0, 0.15, \\
    & & 0, 0, 0)~[$\mathrm{m/Ns^2}$]\\
    
    $I$ & Moments of Inertia & diag(1.6, 0.72, 0.32, \\
    & & 0.3, 0.3, 0.04)~[$\mathrm{kgm^2}$]\\
    
    $g$ & Cutoff freq. of Deriv. filter &  10~[Hz]\\

    $T_{\mathrm{s}}$ &Sample time of the controller &  0.001~[s]\\
    \hline
   \end{tabular}
\end{table}

\end{document}